\documentclass[11pt]{article}

\usepackage[a4paper,margin=1in]{geometry}
\usepackage[T1]{fontenc}
\usepackage[utf8]{inputenc}   % keep for pdfLaTeX compatibility
\usepackage{lmodern}          % fixes many font-shape warnings
\usepackage{microtype}

\usepackage{amsmath,amssymb,amsthm,mathtools}

\usepackage{graphicx}
\usepackage{booktabs}
\usepackage{tabularx}
\usepackage{longtable}
\usepackage{multirow}
\usepackage{siunitx}

\usepackage{float}

\usepackage[ruled,vlined]{algorithm2e}

\usepackage[numbers,sort&compress]{natbib}

\usepackage[hidelinks]{hyperref}
\usepackage{caption}
\usepackage{graphicx}
\usepackage{subcaption}
\usepackage{tikz}
\usetikzlibrary{positioning}
\usepackage{pgfplots}
\pgfplotsset{compat=1.18}

\usepackage[hidelinks]{hyperref}

\usepackage{tabularx}
\usepackage{caption}
\usepackage{tikz}
\usetikzlibrary{positioning}
\usepackage{pgfplots}
\pgfplotsset{compat=1.18}
\usepackage{booktabs}
\usepackage{longtable}
\usepackage{siunitx} 
\usepackage{multirow}
\usepackage{graphicx}
\usepackage{hyperref}

\usepackage{mathtools}
\DeclarePairedDelimiter{\abs}{\lvert}{\rvert}

\title{Brownian ReLU(Br-ReLU): A New Activation Function for a Long-Short Term Memory (LSTM) Network}
\newcommand{\KNUSTaffil}{Department of Statistics and Actuarial Science, Kwame Nkrumah University of Science and Technology, Kumasi, Ghana}
\newcommand{\TMUaffil}{Department of Mathematics, Toronto Metropolitan University, Toronto, Canada}

\author{%
	George Awiakye-Marfo\thanks{\KNUSTaffil} \and
	Elijah Agbosu\footnotemark[1] \and
	Victoria Mawuena Barns\footnotemark[1] \and
	Samuel Asante Gyamerah\thanks{\TMUaffil}%
}
\date{} % leave empty for arXiv style

\begin{document}
	\maketitle
	
	\begin{abstract}
		\noindent
		Deep learning models are effective for sequential data modeling, yet commonly used activation functions such as ReLU, LeakyReLU, and PReLU often exhibit gradient instability when applied to noisy, non-stationary financial time series. This study introduces BrownianReLU, a stochastic activation function induced by Brownian motion that enhances gradient propagation and learning stability in Long Short-Term Memory (LSTM) networks. Using Monte Carlo simulation, BrownianReLU provides a smooth, adaptive response for negative inputs, mitigating the dying ReLU problem. The proposed activation is evaluated on financial time series from Apple, GCB, and the S\&P 500, as well as LendingClub loan data for classification. Results show consistently lower Mean Squared Error and higher $R^2$ values, indicating improved predictive accuracy and generalization. Although ROC–AUC metric is limited in classification tasks, activation choice significantly affects the trade-off between accuracy and sensitivity, with Brownian ReLU and the selected activation functions  yielding practically meaningful performance.
	\end{abstract}

	\section{Introduction}
	%\label{sec:sample1}
	
	In recent years, deep neural networks (DNNs) have revolutionized the approach to solving complex tasks across diverse fields, including pattern recognition, natural language processing, information retrieval, recommendation systems, medical diagnostics, and financial forecasting. Central to the architecture of neural networks, such as convolutional neural networks (CNNs) and recurrent neural networks (RNNs), is the activation function (AF). This critical component introduces nonlinearity, neural networks to capture and represent intricate relationships in data, which is essential for their success in tackling such multifaceted problems. Activation functions are essential in neural networks for learning and interpreting complex, non-linear relationships between input data and the target output. They enable the network to model intricate functional mappings, making it possible to solve sophisticated problems effectively. Without activation functions, the network would only handle simple linear transformations, significantly limiting its utility \cite{dubey2022activation}. Over the past years, numerous activation functions have been proposed, varying in computational complexity and performance. The famous nonlinear monotonic activation function usually include the sigmoid function of \cite{goodfellow2016deep}, this was one of the earliest nonlinear activation functions used in deep learning, mapping all input values to the range (0, 1).  The Tanh function \cite{kalman1992tanh} resolved the non-zero-centered output issue of the sigmoid function but, like sigmoid, suffers from vanishing gradients and saturation, limiting its effectiveness in training deep neural networks. Several
	improvements have been proposed based on the  Sigmoid and Tanh family of activation functions, notably the shifted and scaled sigmoid as used in \cite{arai2018spin}, the scaled hyperbolic tangent activation function of \cite{lecun1998gradient}, the sigmoid-algebraic  and triple - state sigmoid activation functions of \cite{koccak2021new}, the penalised hyperbolic tangent function of \cite{xu2016revise} . Other examples include linearly scaled hyperbolic tangent function of \cite{roy2022lisht}, the sigmoid and tanh combination of \cite{vergara2020study}, the soft-root-sigmoid of \cite{li2020soft}, the sigmoid - Gumbel activation function of \cite{kaytan2022sigmoid} among others.   These extensions aimed to mitigate the vanishing gradient problem but most of them  only provided partial improvements rather than complete solutions.
	However, the introduction of the ReLU activation function by \cite{nair2010rectified}  significantly mitigated this issue due to its simplicity and superior performance, leading to its widespread adoption in deep learning applications 
	as evidenced in various studies e.g . \cite{krizhevsky2012imagenet} and \cite{raj2023improved}. Nonetheless, ReLU had its own limitations, the non-differentiability at zero, unbounded outputs, and the ``dying ReLU" issue, where neurons cease to learn. To overcome these challenges, researchers have developed several modified versions of ReLU. The Leaky ReLu for instance addresses the issue of ``dying neurons" in standard ReLU by allowing a small, nonzero gradient for negative inputs instead of outputting zero. This ``leak" ensures that neurons remain active and continue to learn  for negative inputs, \cite{maas2013rectifier}. Another variant of the ReLU is the  Randomised Leaky ReLU of \cite{xu2015empirical} where the slope of the negative input region is chosen randomly during training. However, the randomised Leaky ReLU (LReLU) struggles with selecting the optimal slope for negative inputs, which can vary depending on the specific problem which  the  Parametric ReLU (PReLU)  of \cite{he2015delving} improves by making the slope for negative inputs a trainable parameter, enabling the model to adaptively learn the best slope during training. Other functions include the bounded ReLU (BReLU) of \cite{liew2016bounded} which is a modified version of ReLU that restricts the output range to address potential instability caused by ReLU's unbounded outputs, %\cite{nair2010rectified},
	the Elastic ReLU and Elastic parametric ReLU of \cite{jiang2018deep}, the softplus of \cite{dugas2000incorporating} as well as ELU and its variants, \cite{clevert2015fast}. %bility during training.  
	The mish and swish  of \cite{misra2019mish} and  \cite{ramachandran2017searching} respectively were smooth, non-monotonic and differentiable activation functions that incorporates a learnable parameter. This parameter enables the model to automatically optimize and adapt the activation function to the specific task during training, offering flexibility and improving performance. The noisy activation function introduced by  \cite{gulcehre2016noisy} % introduced a noisy activation function that incorporates structured, bounded noise into the activation process. 
	also enhances the optimizer's ability to explore the parameter space more effectively and accelerates learning by adding stochastic samples from a normal distribution to the activations, these models mimics the uncertainty and stochastic behavior seen in neural activity,  For an indepth discussion on activation functions see \cite{hammad2024deep, kunc2024three}.  
	On the other hand LSTM introduced by \cite{hochreiter1997long} with its gating mechanisms network was designed to address the key limitation of vanishing gradient  in earlier Recurrent Neural Networks (RNNs).It uses a unique architecture with memory cells and gating mechanisms (input, forget, and output gates) that allow the network to retain important information over long periods and selectively forget irrelevant data and effective for remembering long-term dependencies.  %This makes LSTMs highly effective in tasks that require remembering long-term dependencies, such as:   
	Although LSTMs somewhat solved the vanishing gradient problem and made it possible to learn and retain long-term dependencies, enabling deep learning models to perform better on sequential tasks.  Improved activation functions in LSTM models are sought after to enhance the model's learning capacity, performance, and stability, addressing various challenges in training and optimizing LSTMs, especially in deep learning tasks involving sequential data.This study introduces Brownian ReLU (Br-ReLU), a modified ReLU activation function incorporating stochastic elements inspired by Brownian motion. 
	
	The introduction of Brownian noise into the LSTM network’s activation function is to facilitate smoother gradient flow, potentially improving learning dynamics. 
	This idea combines the deterministic form of ReLU with randomness from Brownian motion, potentially improving model robustness. %%addressing the dying ReLU problem, in an LSTM architecture. 
	We hypothesise that enhanced gating mechanisms using the Br-ReLu would improve the network's ability to model uncertainty in data, particularly in domains like financial forecasting, climate modeling where stochasticity plays a critical role. We test this new formulation with the closing prices of S\&P 500, Apple, Ghana Commercial Bank (GCB) stock data  and an LSTM-based classifier for the LendingClub loan data for status prediction. This paper is structured as follows: Section 2 discusses the related works and  the model formulation. The results and discussion are presented in Section 3, and we finally conclude in Section 4.

	\section{Activation Functions}
	We review the relative nonlinear activation functions of Rectified Linear Unit (ReLU), Leaky  ReLU, and the parametric ReLU as well as introduce the proposed Brownian ReLU.  
	\subsection{ReLU, LReLU, Parametric ReLU and  Noisy Activation Functions}
	\vspace{0.2cm}
	\begin{itemize}
		\item[(a)] 
		The ReLU activation function of \cite{nair2010rectified} is:
		\[
		f(x) = \max(0, x)
		\]
		If \(x > 0\), \(f(x) = x\) (the function outputs the input directly).
		If \(x \leq 0\), \(f(x) = 0\) (the function outputs zero).
		\item[(b)] The Leaky ReLU activation function is given as:
		\[
		f(x) = 
		\begin{cases} 
			x & \text{if } x > 0 \\
			\alpha x & \text{if } x \leq 0 
		\end{cases}
		\]
		Here:
		\(x\) is the input,  \(\alpha\) is a small positive constant (e.g., \(0.01\)) that determines the slope for \(x \leq 0\).
		The Leaky ReLU addresses the ``dying ReLU" problem by allowing a small, non-zero gradient for negative values of \(x\), ensuring neurons do not become inactive, \cite{maas2013rectifier}
		\item[(c)] The parametric ReLU (PReLU) activation function is a variation of the ReLU activation function where the slope for negative inputs is learned during training. It is given as :
		\[
		f(x) =
		\begin{cases} 
			x & \text{if } x > 0, \\
			\alpha x & \text{if } x \leq 0,
		\end{cases}
		\]
		where: \\
		\(x\) is the input to the activation function. \(\alpha\) is a learnable parameter that adjusts the slope for negative inputs. It is initialized and updated during training using GD or other optimization algorithms, allowing the network to learn an optimal value for the negative slope. This flexibility allows PReLU to adapt better to specific tasks, potentially improving performance by addressing the ``dying ReLU" problem, \cite{he2015delving}.
		%\end{itemize}
		\item[(d)] The Gaussian Error Linear Unit (GELU) activation function of \cite{hendrycks2016gaussian} weights inputs by their value rather than gating them as ReLUs do. It is given by:
		\[
		\text{GELU}(x) = x \cdot \Phi(x),
		\]
		where \( \Phi(x) \) is the cumulative distribution function (CDF) of the standard normal distribution:
		\[
		\Phi(x) = \frac{1}{2} \left[ 1 + \text{erf}\left(\frac{x}{\sqrt{2}}\right) \right],
		\]
		and \( \text{erf}(x) \) is the error function. This approximation is computationally efficient and commonly used in practice. 
	\end{itemize}

	\subsection{Proposed Brownian ReLU (Br-ReLU)  Activation Function}
\begin{figure}[H]
		\centering
		\includegraphics[width= 0.65\linewidth]{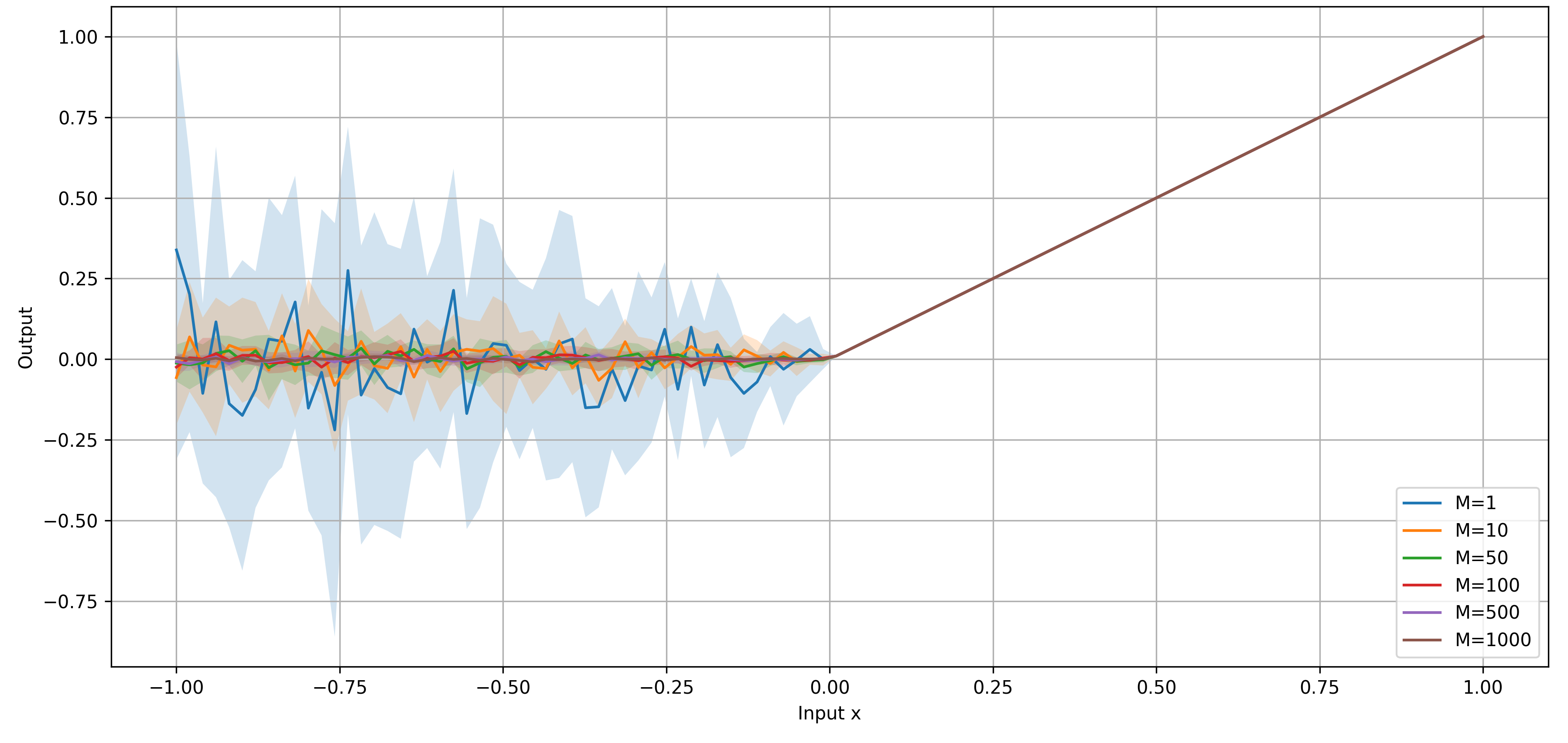}
		\caption{Illustrative Monte Carlo paths for Br-ReLU Activation Function}
		\label{fig:enter-label_1}
	\end{figure}

	The BrownianReLU activation function leverages the stochastic properties of Brownian motion to introduce adaptive nonlinearity into neural networks and make it more adaptive to actual data. However, since standard Brownian motion is  not defined over non-negative inputs, its direct applicability to negative inputs is limited, we therefore the employ the  \textit{symmetry principle} of Brownian motion. \cite{karatzas2014brownian}.
	This principle enables the extension of Brownian motion behaviour to negative values through reflection or symmetry.We give the activation function in equation \eqref{gh2} below; 
	\begin{equation}\label{gh2}
		f(x) =
		\begin{cases}
			x & \text{for } x > 0, \\
			-\alpha  B(|x|) & \text{for } x \leq 0,
		\end{cases}
	\end{equation}
	where  $B (|x|) \sim \mathcal{N}(0, \abs{x})$ and $\alpha$ is a learnable parameter controlling the negative slope, consistent with that of the PReLU activation function. 
	\cite{he2015delving}. BrownianReLU returns the identity for positive inputs, while for negative inputs it produces a stochastic path, generated through Monte Carlo simulations, as shown in equation \eqref{gh10} and illustrated in Figure \eqref{fig:enter-label_1}.
	\begin{equation}\label{gh10}
		f(x) =
		\begin{cases}
			x  & \text{for } x > 0, \\
			-\alpha \cdot \dfrac{1}{M} \sum\limits_{i=1}^M  B^{k}(|x_i|) & \text{for } x \leq 0,
		\end{cases}
	\end{equation}
	where $B(\abs{x}) \sim \mathcal{N}(0, \abs{x})$ and  $M$ is the number of sample paths. Figure \eqref{fig:enter-label_1}, shows the Brownian ReLU activation for some  $\alpha$ and different values of $M$.
	
	It demonstrates the limiting behavior of the Monte Carlo mean path for negative inputs as the number of simulated sample paths increases from 1 to 1000. As M grows, the mean path converges with reduced variance and greater smoothness, consistent with the law of large numbers.

	\subsubsection{Alpha Adaptation}
	The adaptation of the parameter \( \alpha \) in Br-ReLU is fundamental in regulating the behavior of the activation function, particularly within the negative input domain.  Similarly, we explore the training process of $\alpha$ in \cite{he2015delving} in the proposed Br-ReLU. This enables the model to adjust the degree of stochastic perturbation applied to negative inputs based on the data distribution. %and task complexity. This dynamic adaptation allows the model to effectively manage the integration of noise and uncertainty.
	\begin{figure}[h!]
		\centering
		\begin{minipage}[t]{0.49\textwidth}
			\includegraphics[width=1.1\textwidth]{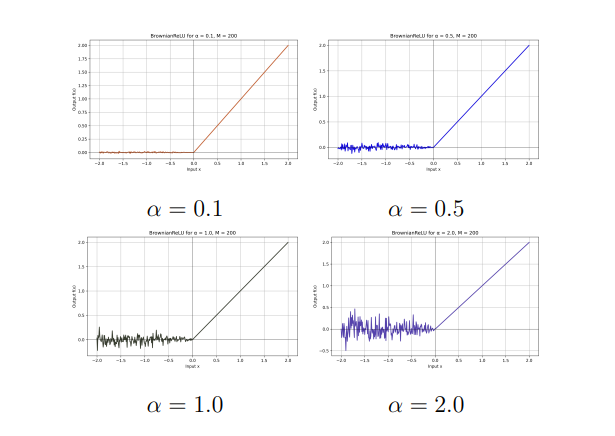}
			\caption{Varying values of $\alpha$ at M=200}
			\label{fig:sub0} 
		\end{minipage}
		\begin{minipage}[t]{0.49\textwidth}
			\includegraphics[width=1.1\textwidth]{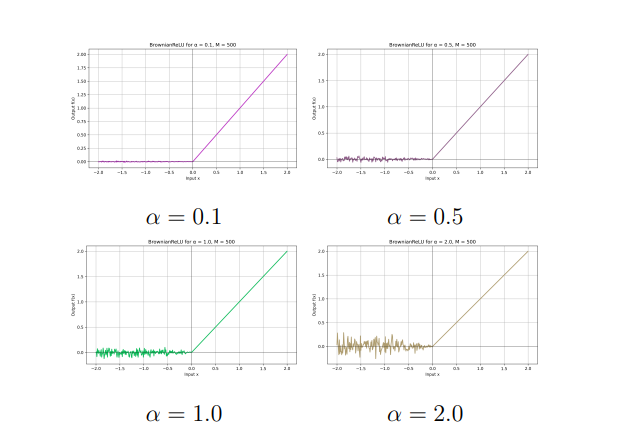}
			\caption{Varying values of $\alpha$ at M=500}
			\label{fig:sub1}
		\end{minipage} 
		\begin{minipage}[t]{0.49\textwidth}
			\includegraphics[width=1.1\textwidth]{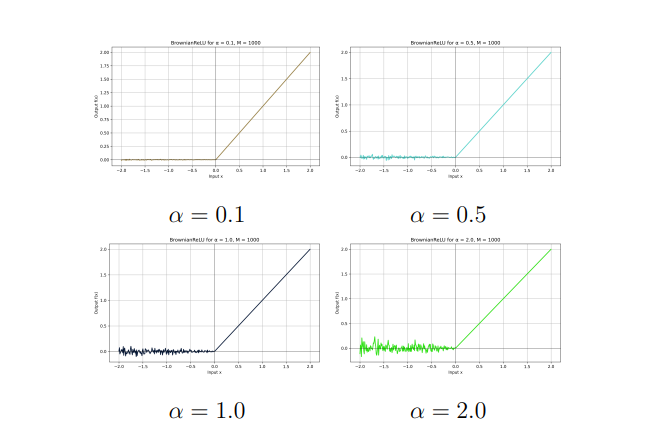}
			\caption{Varying values of $\alpha$ at M=1000}
			\label{fig:sub2}
		\end{minipage}
		\begin{minipage}[t]{0.49\textwidth}
			\includegraphics[width=1.1\textwidth]{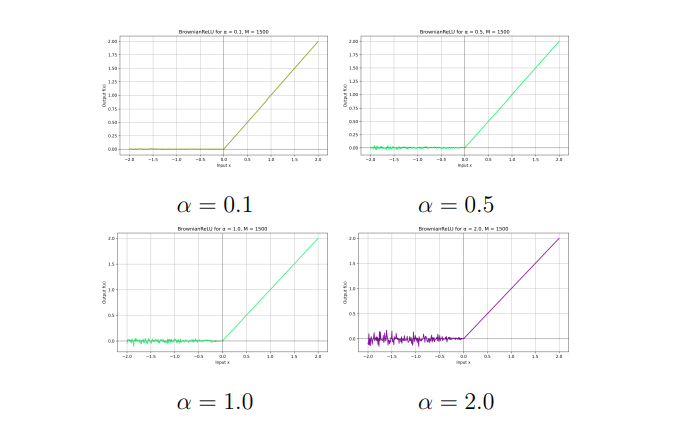}
			\caption{Varying values of $\alpha$ at M=1500}
			\label{fig:sub3}
		\end{minipage}
		\caption{Figure showing subfigures 2, 3, 4 and 5}
		\label{fig:main}
	\end{figure}
	Figure \eqref{fig:main} illustrates the effect of varying alpha with increasing sample sizes M. For values of 
	$M=200, 500$, the Monte Carlo mean path remains relatively noisy, whereas larger M values (1000 and 1500) produce smoother and more stable paths. Across all cases, alpha controls the scaling on the negative input side, with higher values amplifying stochastic fluctuations and smaller values keeping the path closer to zero.  We note that when $\alpha = 0$ we obtain the ReLU function.

	\subsubsection{Gradient computation and parameter optimisation for BrownianReLU}
	Consider the activation in \eqref{gh1}
	\begin{equation}\label{gh1}
		f(x;\alpha)=
		\begin{cases}
			x  & \text{for } x > 0, \\
			-\alpha \cdot \dfrac{1}{M} \sum\limits_{k=1}^M  B^{k}(|x_i|) & \text{for } x \leq 0,
		\end{cases}
	\end{equation}
	where for each input \(x\) the values \(B^{(k)}(|x|)\), \(k=1,\dots,M\), are independent draws with
	\(B^{(k)}(|x|)\sim\mathcal{N}(0,|x|)\) and \(\alpha\in\mathbb{R}\) is a learnable parameter. %\cite{he2015delving}.
	Let the model output for sample \(i\) be \(y_i=f(x_i;\alpha)\) 
	and let the Monte Carlo average of the sample be
	$$
	b_i \;:=\; \frac{1}{M}\sum_{k=1}^M B^{(k)}(|x_i|) \,,
	\hspace{0.5cm}
	\text{so that}
	\hspace{0.5cm}
	f(x_i;\alpha)=
	\begin{cases}
		x, & x>0,\\[6pt]
		-\,\alpha\, b_i, & x_i\le 0.
	\end{cases}
	$$
	Again let \(\delta_i=\partial L_i/\partial y_i\), be the gradient at the activation, where \(L_i\) is the loss for sample \(i\) 
	Then
	\[
	\frac{\partial f(x_i;\alpha)}{\partial \alpha}
	= \begin{cases}
		0, & x >0,\\[6pt]
		-\,b_i, & x_i\le 0.
	\end{cases}
	\]
	Hence, the contribution per-sample to the gradient of the loss w.r.t.\ \(\alpha\) is
	\[
	\frac{\partial L_i}{\partial \alpha}
	= \delta_i\frac{\partial y_i}{\partial \alpha}
	= \begin{cases}
		0, & x_i>0,\\[6pt]
		-\,\delta_i\,b_i, & x_i\le 0.
	\end{cases}
	\]
	For a minibatch \(\mathcal B\) the gradient estimator is the sum over the batch:
	$$
	\frac{\partial L_{\mathcal B}}{\partial \alpha}
	= -\sum_{i\in\mathcal B}\delta_i\,\mathbf{1}_{\{x_i\le 0\}}\,b_i
	= -\sum_{i\in\mathcal B}\delta_i\,\mathbf{1}_{\{x_i\le 0\}}\;\frac{1}{M}\sum_{k=1}^M B^{(k)}(|x_i|).
	$$
	With a learning rate $\eta$ a simple SGD step can be obatined as;
	$$
	\alpha \leftarrow \alpha + \eta \sum_{i\in\mathcal B}\delta_i\,\mathbf{1}_{\{x_i\le 0\}}\,b_i
	\;=\;
	\alpha + \eta \sum_{i\in\mathcal B}\delta_i\,\mathbf{1}_{\{x_i\le 0\}}\left(\frac{1}{M}\sum_{k=1}^M B^{(k)}(|x_i|)\right).
	$$

	\subsubsection{Training Procedure with BrownianReLU Activation}
	\begin{algorithm}[H]
		\caption{Training Procedure with BrownianReLU Activation}
		\KwIn{Training data $\{(x_i, t_i)\}$, learning rate $\eta$, number of Monte Carlo samples $M$}
		\KwOut{Updated model parameters (including $\alpha$)}
		
		for each minibatch $\mathcal{B}$ do\\
		\hspace{1em}for each input $x_i$ in $\mathcal{B}$ do\\
		\hspace{2em}\textbf{if} $x_i > 0$ \textbf{then}\\
		\hspace{3em}$y_i \leftarrow x_i$\\
		\hspace{2em}\textbf{else}\\
		\hspace{3em}\textbf{for} $k = 1$ to $M$ \textbf{do}\\
		\hspace{4em}Sample $B^{(k)}(|x_i|) \sim \mathcal{N}(0, |x_i|)$\\
		\hspace{3em}\textbf{end for}\\
		\hspace{3em}$\bar{B}(|x_i|) \leftarrow \frac{1}{M} \sum_{k=1}^{M} B^{(k)}(|x_i|)$\\
		\hspace{3em}$y_i \leftarrow -\alpha \cdot \bar{B}(|x_i|)$\\
		\hspace{2em}\textbf{end if}\\[0.5em]
		\hspace{1em}\textbf{end for}\\[0.5em]
		\hspace{1em}Compute minibatch loss:
		$L_{\mathcal{B}} = \sum_{i \in \mathcal{B}} L(y_i, t_i)$\\
		\hspace{1em}Backpropagate to compute:
		$\delta_i = \frac{\partial L_i}{\partial y_i}$\\
		\hspace{1em}Compute gradient w.r.t. $\alpha$:
		$\frac{\partial L_{\mathcal{B}}}{\partial \alpha}
		= -\sum_{i \in \mathcal{B}} \delta_i \cdot \mathbf{1}_{\{x_i \leq 0\}} \, \bar{B}(|x_i|)$\\
		\hspace{1em}Update $\alpha$:
		$\alpha \leftarrow \alpha - \eta \, \frac{\partial L_{\mathcal{B}}}{\partial \alpha}$\\
		\hspace{1em}Update other network parameters using standard SGD\\
		\textbf{end for}
	\end{algorithm}

	\subsection{LSTM System Model}
	We apply the proposed activation on a basic LSTM architecture that consists of  three main components: an input layer,  a hidden layer, and an output layer. %The hidden layer contains single-cell blocks with recurrently connected units. 
	At each time step \( t \), the input vector \( x_t \) is fed into the network, the components of the block governed by the following equations.
	\begin{itemize}
		\item[(a)] The forget gate determines what information should be discarded from the cell state.
		\[
		f_t = \sigma(W_f x_t + U_f h_{t-1} + b_f)
		\]
		
		\item[(b)] The input gate controls what information to add to the cell state.
		\[
		i_t = \sigma(W_i x_t + U_i h_{t-1} + b_i)
		\]
		\item[(c)] The input gate is complemented by a candidate cell state, given as:
		\[
		\tilde{C}_t = \tanh(W_c x_t + U_c h_{t-1} + b_c)
		\]
		\item[(d)] The cell state \( C_t \) is updated using the forget gate, input gate, and candidate cell state:
		\[
		C_t = f_t \odot C_{t-1} + i_t \odot \tilde{C}_t
		\]
		Where \( \odot \) represents element-wise multiplication.
		\item[(e)] The output gate determines what information to output from the cell.
		\[
		o_t = \sigma(W_o x_t + U_o h_{t-1} + b_o)
		\]
		\item[(f)] Finally, the hidden state \( h_t \) is updated as:
		\[
		h_t = o_t \odot \tanh(C_t)
		\]
	\end{itemize}
	where; $f_t$, $i_t$ and $o_t$ are the  functions of the forget, input and output gates respectively. \( W_f \) and \( U_f \) are weights applied to the current input \( x_t \) and previous hidden state \( h_{t-1} \), respectively,   \( b_f \) is the bias vector and \( \sigma \) is the sigmoid activation  function.  \( \tilde{C}_t \) is the candidate cell state,  \( W_i, W_c, U_i, U_c \), and \( b_i, b_c \) are weights and biases,  \( \tanh \) is the hyperbolic tangent activation function. 
	In this study the proposed activation function is applied to the cell state  and the hidden state, these  are  key components in the model \cite{hochreiter1997long}. %goodfellow2016deep}.
	The equation of the Cell State and the hidden state is now updated to;
	\begin{align}
	\tilde{C}_t = \text{Br-ReLU}(W_{xc} X_t + W_{hc} h_{t-1} + b_c)
	\label{eqn 3}
	\end{align}
	and
	\begin{align}
	h_t = o_t \text{Br-ReLU}(C_t)
	\label{eqn 4}
	\end{align}

	\section{Numerical Results}
	Here, we perform a  comparison among the various activation functions for the various different datasets notably the Ghana commercial Bank (GCB) stock data, Apple and S\&P 500.

	\subsection{Descriptive Statistics}
	Table \ref{tab:convergence_1} presents the mean and variance values for the Apple, GCB, and S\&P 500 stock datasets.  Amongst the three, the S\&P 500 shows the highest mean (0.3545) and variance (0.0626), indicating relatively larger average returns and greater volatility. Apple follows closely with a mean of 0.3267 and variance of 0.0576, showing moderate fluctuations in stock movements. GCB, on the other hand, records the lowest variance (0.0326), suggesting more stable returns. 
	
	\begin{table}[H] 
		\caption{Variance of Apple, GCB, and S\&P500 stock dataset} \centering \begin{tabular}{ccccccc} \toprule No. & Data & Mean & Variance %& Skewness
		\\ \midrule 1 & Apple & 0.326708 & 0.057563 %&  0.834408
		\\ 2 & S\&P500 & 0.354458 & 0.062592 %& 0.517173 
		\\ 3 & GCB & 0.341235 & 0.032637 %& -0.154725
		\\ \bottomrule 
		\end{tabular} \label{tab:convergence_1} \end{table}

\subsection{Sensitivity Analysis}
		\begin{longtable}{cccS[table-format=2.4]S[table-format=1.6]S[table-format=1.6]S[table-format=1.6]S[table-format=1.6]c}
	\caption{Sensitivity analysis of BrownianReLU sample paths across datasets} 
	\label{tab:sensitivity_analysis} \\
	\toprule
	Data & {Evaluation Matrix} & \multicolumn{5}{c}{Evaluation at Different \( M \)} \\
	\cmidrule(lr){3-7}
	& & {$M=500$} & & {$M=1000$} & & {$M=1500$} \\
	\midrule
	\endfirsthead
	% Header for subsequent pages
	\multicolumn{9}{c}%
	{{\bfseries Table \thetable{} -- continued from previous page}} \\
	\toprule
	Data & {Evaluation Matrix} & \multicolumn{5}{c}{Evaluation at Different \( M \)} \\
	\cmidrule(lr){3-7}
	& &  {$M=500$} & & {$M=1000$} & & {$M=1500$}  \\
	\midrule
	\endhead
	% Footer for pages except the last
	\midrule
	\multicolumn{9}{r}{{Continued on next page}} \\
	\endfoot
	% Footer for the last page
	\bottomrule
	\endlastfoot
	% --------- DATA ROWS -----------
	& MSE & 0.002122 & & 0.003162 & & 0.002035  \\
	Apple & $R^2$(Test) & 0.9355 & & 0.9036 & & 0.9381 \\
	& Epoch & 48 & & 46 & & 49  \\
	\cmidrule(lr){1-7}
	& MSE & 0.000284 & & 0.000275 & & 0.000276  \\
	GCB & $R^2$(Test) & 0.9864 & & 0.9869 & & 0.9868 \\
	& Epoch & 44 & & 44 & & 36  \\
	\cmidrule(lr){1-7}
	& MSE & 0.00038 & & 0.000232 & & 0.000585  \\
	SP500 & $R^2$(Test) & 0.9829 & & 0.9896 & & 0.9736 \\
	& Epoch & 50 & & 50 & & 50  \\
	\label{sense}
	\end{longtable}
	Table \ref{tab:sensitivity_analysis} shows the sensitivity analysis of the proposed BrownianReLU activation function across the three datasets: Apple, GCB, and S\&P500. The evaluation was based on the mean squared error (MSE), the coefficient of determination on the test set ($R^2$), and the number of epochs required for training. Performance was assessed under three different sample path sizes, namely $M=500$, $M=1000$, and $M=1500$.For the Apple dataset, MSE values range between 0.002122 and 0.002035, with $R^2$ values varying from 0.9036 to 0.9381, indicating moderate sensitivity to the choice of $M$. The GCB dataset exhibits highly stable behavior, with consistently low MSE values (approximately 0.00028) and  $R^2$ values near 0.987 across all values of $M$. On the otherhad, the SP500 dataset shows greater variation: the lowest MSE (0.000232) and highest $R^2$ (0.9896) occur at $M=1000$, while performance declines at $M=1500$ with MSE increasing to 0.000585 and $R^2$ falling to 0.9736 with varying numbers of epochs also differs across datasets. For Apple, convergence occurs between 46 and 49 epochs, while GCB converges within 36–44 epochs. The SP500 dataset, however, consistently requires 50 epochs across all settings. The results suggest that the choice of sample paths ($M$) can influence both accuracy and training efficiency, particularly for datasets with higher variability such as S\&P500. $M=1000$ appears to provide a favorable balance between error minimization and performance generalisation.

	\subsection{Evaluating BrownianReLU Against Standard Activation Functions on Multiple Datasets}
	\begin{table}[h!]
	\caption{Comparison of Standard activation functions with Brownian RELU  using the Apple dataset}
	\centering
	\begin{tabular}{ccccccc}
		\toprule
		No. & Activation Function & MSE & $R^2$(Train) & $R^2$(Test) & Epoch of convergence \\
		\midrule
		1 & BrownianReLU & 0.002035 & 0.9903 & 0.9381 & 49\\
		2 & LeakyReLU & 0.160918 & 0.9543 & 0.5546& 48\\
		3 & PReLU & 0.043347 & 0.9904 & 0.7425 & 13\\
		4 & ReLU & 0.005931 & 0.9881 & %0.8197 epoch 18
		0.3164& 36\\
		5 & Tanh & 0.004110 & 0.9592 & %0.8750 
		0.7297& 3 \\
		\bottomrule
	\end{tabular}
	\label{tab:convergence_5}
	\end{table}
	Table \ref{tab:convergence_5} compares the proposed BrownianReLU activation function with the following activation function LeakyReLU, PReLU, ReLU, and Tanh using the Apple dataset. BrownianReLU activation function achieved the lowest MSE (0.002035) and the highest test $R^2$ (0.9381), indicating superior predictive performance. While PReLU and ReLU performed well during training, their lower test $R^2$ values suggest weaker generalization. LeakyReLU produced the poorest results with a negative test $R^2$, whereas Tanh converged quickly with moderate accuracy. 
	
	\begin{table}[H]
	\caption{Comparison of Standard activation functions with Brownian RELU  using the GCB dataset}
	\centering
	\begin{tabular}{ccccccc}
		\toprule
		No. & Activation Function &MSE & $R^2$(Train) & $R^2$(Test) & Epoch of convergence \\
		\midrule
		1 & BrownianReLU & 0.000275 & 0.9982 & 0.9869 & 44\\
		2 & LeakyReLU & 0.000303 & 0.9980 & 0.9855 & 46\\
		3 & PReLU & 0.000291 & 0.9985 & 0.9861 & 44\\
		4 & ReLU & 0.000288 & 0.9985 & 0.9862 & 48\\
		5 & Tanh & 0.000333 & 0.9980 & 0.9841 & 48 \\
		6 & GELU & 0.000387 & 0.9976 & 0.9815 & 50\\
		\bottomrule
	\end{tabular}
	\label{tab:convergence_6}
	\end{table}
	The results in Table \ref{tab:convergence_6} shows  the performance metrics across all models. The result show  generally consistent result across all metrics, reflecting the inherent stability of the dataset. Among the compared  functions, BrownianReLU recorded the lowest MSE (0.000275) and the highest test $R^2$ (0.9869), demonstrating a slight yet meaningful improvement in predictive accuracy. Although ReLU and PReLU, Tanh, and GELU  exhibited comparable performance, the GELU activation function produced the highest MSE and lowest test $R^2$, indicating relatively weaker generalization. Overall, the BrownianReLU showed strong accuracy and stable convergence and hence its effectiveness on the GCB dataset.
	\begin{table}[h!]
	\caption{Comparison of the standard  Activation functions to the proposed activation function using the S\&P500 dataset}
	\centering
	\begin{tabular}{ccccccc}
		\toprule
		No. & Activation Function &MSE & $R^2$(Train) & $R^2$(Test) & Epoch of convergence \\
		\midrule
		1 & BrownianReLU & 0.000242 & 0.9973 & 0.9891 & 48\\
		2 & LeakyReLU & 0.011240 & 0.9380 & 0.4917 & 2\\
		3 & PReLU & 0.002399 & 0.9965 & 0.8915 & 42\\
		4 & ReLU & 0.003820 & 0.9964 & 0.8272 & 24 \\
		5 & Tanh & 0.000255 & 0.9972 & 0.9841 & 50 \\
		6 & GELU & 0.012362 & 0.9931 & 0.4426 & 37 \\
		\bottomrule
	\end{tabular}
	\label{tab:convergence_7}
	\end{table}
	In Table \ref{tab:convergence_7} among all models, BrownianReLU achieved the lowest MSE (0.000242) and the highest test $R^2$ (0.9891), indicating superior predictive accuracy.  The Tanh activation also performed competitively, with similar MSE and $R^2$ values, though it required slightly more epochs to converge. In contrast, LeakyReLU produced the weakest performance, with a much higher MSE (0.011240) closely followed by GELU  and a test $R^2$ of 0.4917 and 0.4426 respectivley, suggesting relatively poorer adaptability to the dataset. PReLU and ReLU showed reasonable results but lagged behind BrownianReLU in both accuracy and stability.

	\subsection{Visual Predictions of BrownianReLU Across Various Datasets.}
	The following figures illustrate the predictive performance of the proposed BrownianReLU activation function across different financial datasets. Each figure compares the predicted stock price trends of the model with the actual data.

\begin{figure}[t]
	\centering
	
	\begin{subfigure}[b]{0.32\textwidth}
		\centering
		\includegraphics[width=\textwidth]{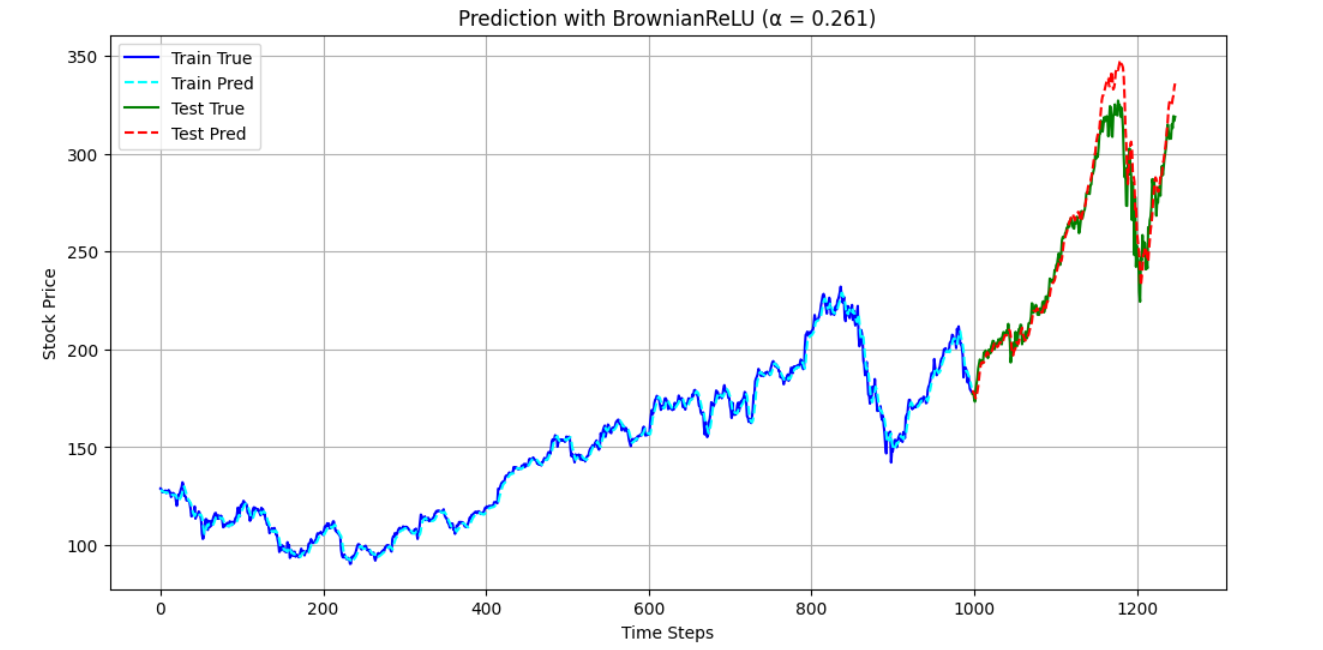}
		\caption{Apple stock prediction}
		\label{fig:apple_prediction}
	\end{subfigure}\hfill
	\begin{subfigure}[b]{0.32\textwidth}
		\centering
		\includegraphics[width=\textwidth]{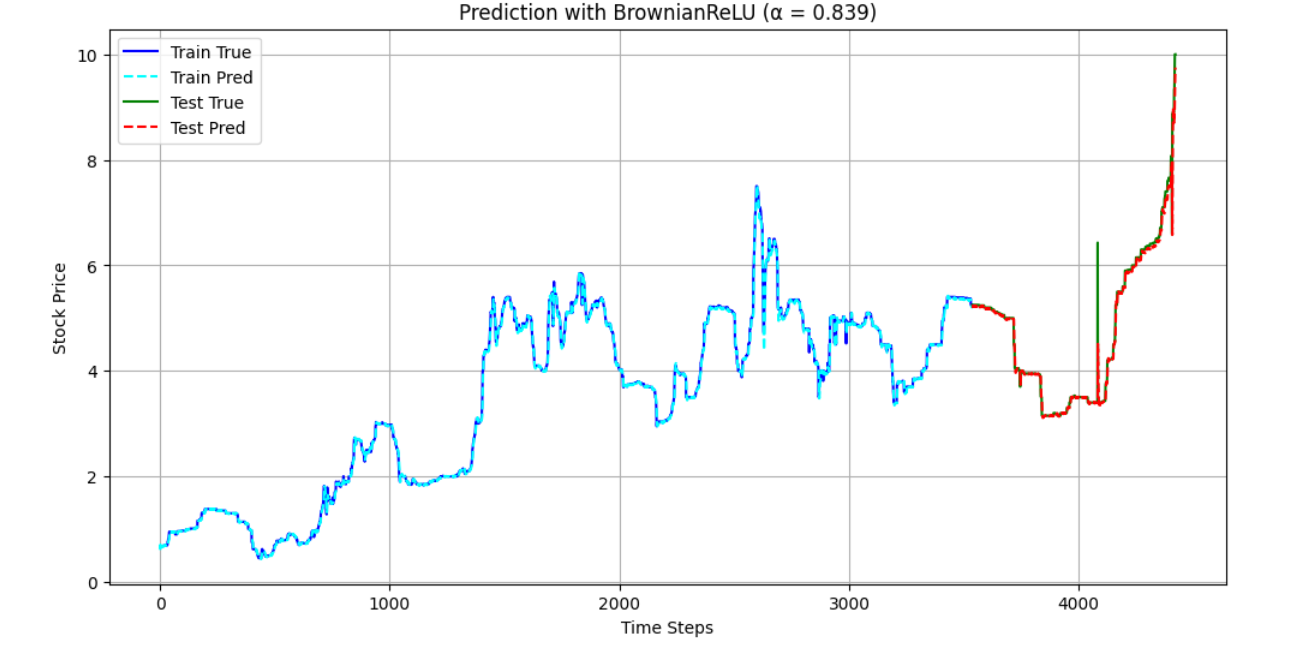}
		\caption{GCB stock prediction}
		\label{fig:gcb_prediction}
	\end{subfigure}\hfill
	\begin{subfigure}[b]{0.32\textwidth}
		\centering
		\includegraphics[width=\textwidth]{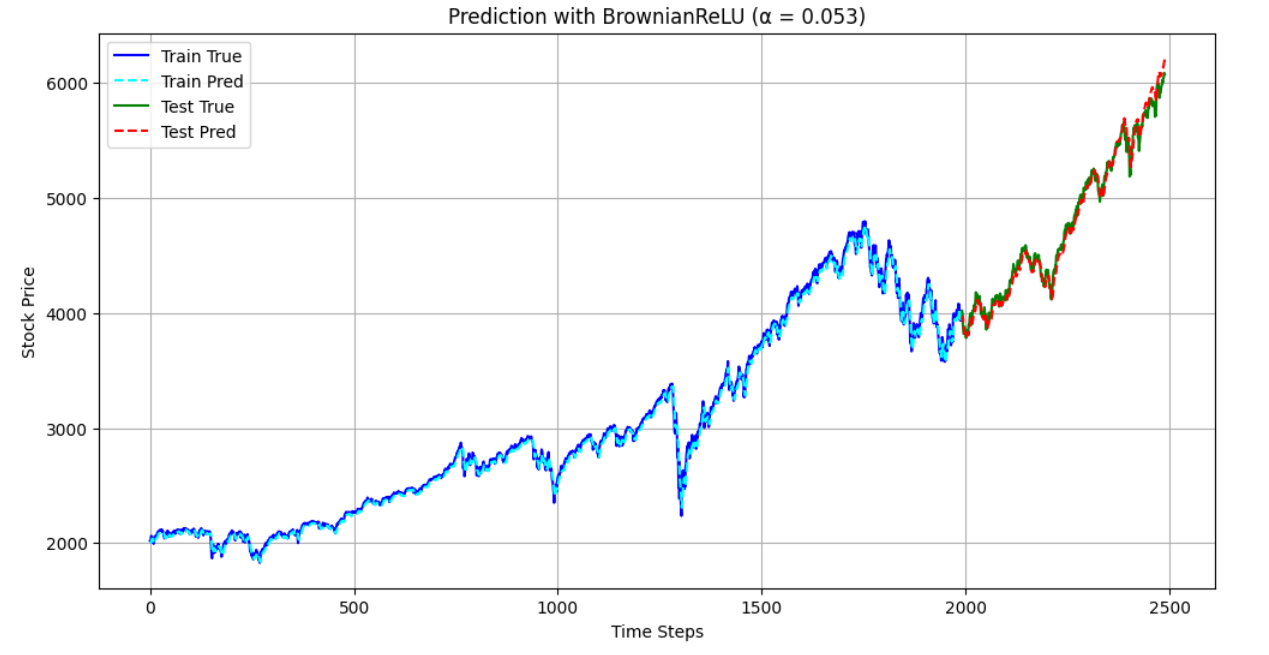}
		\caption{S\&P 500 prediction}
		\label{fig:sp500_prediction}
	\end{subfigure}
	
	\caption{Visual predictions of the BrownianReLU model across the Apple, GCB, and S\&P 500 datasets.}
	\label{fig:combined_predictions}
\end{figure}
	Across all three datasets, the predicted values closely align with the actual stock movements, indicating the model’s ability to capture nonlinear patterns and dependencies effectively.
	For the Apple dataset, the predictions exhibit a smooth fit with slight or minimal  deviations, indicating strong learning stability and generalization. In the GCB dataset, the predictions maintain consistent accuracy, suggesting robustness even in relatively low-volatility financial data. The S\&P 500 predictions display slightly more fluctuations due to the index’s inherent market complexity, however, the model successfully tracks the major trends and fluctuations, see Figure \eqref{fig:combined_predictions}.The results demonstrate that BrownianReLU provides high predictive accuracy across varying market behaviors, validating its effectiveness in handling  volatility and long-term dependencies in financial time series and risk analytics.

	\subsection{Classification Performance Analysis}
	This section evaluates the effect of the various activation functions on the performance of an LSTM-based classifier for loan status prediction. The proposed BrownianReLU is compared with  standard activations shown in Table~\ref{tab:activation_results}. Performance is assessed using Accuracy, Precision, Recall, F1-score, and ROC–AUC. Given the inherent class imbalance in loan payment data, emphasis is placed on Recall, F1-score, and ROC–AUC, which more accurately reflect the model’s ability to detect minority-class outcomes.
	
	\begin{table}[H]
	\centering
	\caption{Classification Performance of LSTM with Different Activation Functions}
	\label{tab:activation_results}
	\begin{tabular}{lccccc}
		\toprule
		\textbf{Activation Function} & \textbf{Accuracy} & \textbf{Precision} & \textbf{Recall} & \textbf{F1-score} & \textbf{ROC--AUC} \\
		\midrule
		BrownianReLU ($\alpha = 0.014$) & 0.7261 & 0.1845 & 0.1033 & 0.1324 & 0.5036 \\
		BrownianReLU ($\alpha = 0.464$) & 0.7250 & 0.2381 & 0.1630 & 0.1935 & 0.5208 \\
		BrownianReLU ($\alpha = 0.480$) & 0.7624 & 0.1667 & 0.0435 & 0.0690 & 0.5118 \\
		BrownianReLU ($\alpha = 0.925$) & 0.6667 & 0.2153 & 0.2446 & 0.2290 & 0.5148 \\
		BrownianReLU ($\alpha = 0.944$) & 0.7802 & 0.1923 & 0.0272 & 0.0476 & 0.5236 \\
		\midrule
		ReLU      & 0.7030 & 0.2278 & 0.1957 & 0.2105 & 0.5218 \\
		LeakyReLU & 0.6634 & 0.2010 & 0.2228 & 0.2113 & 0.5150 \\
		PReLU     & 0.7767 & 0.2683 & 0.0598 & 0.0978 & 0.5297 \\
		tanh      & 0.7305 & 0.1919 & 0.1033 & 0.1343 & 0.5222 \\
		GELU      & 0.6447 & 0.2043 & 0.2609 & 0.2291 & 0.5113 \\
		\bottomrule
	\end{tabular}
	\end{table}
	
	Table~\ref{tab:activation_results} summarizes the classification results obtained across all activation functions. The models achieved moderate predictive performance, with ROC-AUC values ranging approximately between $0.50$ and $0.53$, indicating classification performance marginally better than random guessing. Among the BrownianReLU variants, performance varied with the choice of the parameter $\alpha$. While BrownianReLU with $\alpha = 0.944$ achieves the highest accuracy $(0.7802)$, it exhibits extremely low recall $(0.0272)$ and F1-score $(0.0476)$, suggesting strong bias toward the majority class. Howevr, BrownianReLU with $\alpha = 0.925$ provides a more balanced trade-off, achieving a recall of $0.2446$, an F1-score of $0.2290$, and a ROC--AUC of $0.5148$.Standard activation functions show mixed performance. ReLU demonstrates reasonable accuracy but limited recall. LeakyReLU improves sensitivity to minority-class observations, reflected in higher recall and F1-score. PReLU achieves relatively high precision but suffers from low recall. The tanh activation offers a balanced compromise between precision and recall, while GELU achieves the highest recall $(0.2609)$ and F1-score $(0.2291)$ among all standard activations, indicating superior capability in capturing nonlinear relationships in the data.
	
	\subsubsection{Interpretation and Implications}
	The results highlight that higher accuracy does not necessarily imply better classification performance in imbalanced datasets. Activation functions that improve recall and F1-score, such as GELU and well-tuned BrownianReLU variants, are more suitable for credit risk and loan default prediction tasks, where identifying high-risk cases is of greater importance than overall accuracy.  Although none of the activation functions yields a substantial improvement in ROC--AUC, the choice of activation significantly affects the balance between accuracy and sensitivity. GELU and selected BrownianReLU configurations provide the most practically meaningful results.

	\section{Conclusion}
	This study examined the effectiveness of the proposed BrownianReLU activation function within LSTM frameworks for financial time series forecasting and classification, addressing the limitations of conventional activations in modeling noisy and volatile data. Based on normalized and sequenced datasets from Apple, GCB, and the S\&P 500, the performance of BrownianReLU was evaluated against ReLU, LeakyReLU, PReLU, and Tanh using Mean Squared Error (MSE), coefficient of determination ($R^2$), and epochs to convergence as key metrics.
	The results consistently showed that BrownianReLU achieved lower MSE and higher $R^2$ values across testing datasets, indicating improved forecasting accuracy and stronger generalization. The learnable parameter $\alpha$ allows dynamic adaptation during training, addressing the dying neuron problem and reducing overfitting through controlled randomness. Visual analyses further confirmed smoother learning curves and enhanced trend-capturing ability. Under classification,  although none of the activation functions yielded a substantial improvement in ROC--AUC, the choice of activation significantly affects the balance between accuracy and sensitivity. GELU and selected BrownianReLU configurations provide the most practically meaningful results. The findings demonstrate that BrownianReLU provides a robust and efficient activation mechanism for financial forecasting tasks and classification.

	\section*{Acknowledgements}
This research has been supported in part by the Faculty of Science at Toronto Metropolitan University.
	
	% ---------- Bibliography ----------
%	% Option A (recommended): BibTeX on arXiv
%	\bibliographystyle{plainnat}
%	\bibliography{references} % references.bib
	
	% Option B (if arXiv has trouble): comment out the two lines above and upload main.bbl instead.

\end{document}